\documentclass{article}

\usepackage[preprint, nonatbib]{neurips_2024}

\usepackage{subfigure}
\usepackage{subcaption}
\usepackage[utf8]{inputenc} 
\usepackage[T1]{fontenc}    
\usepackage{hyperref}       
\usepackage{url}            
\usepackage{booktabs}       
\usepackage{amsfonts}       
\usepackage{nicefrac}       
\usepackage{microtype}      
\usepackage{xcolor}         
\usepackage{lineno}
\usepackage{graphicx}
\usepackage{enumitem}
\usepackage{float}
\usepackage{amsmath}
\usepackage{wrapfig}
\usepackage{caption}
\usepackage{enumerate}
\usepackage{amsthm}
\usepackage{multirow}
\usepackage{algorithm}
\usepackage{algorithmic}

\usepackage{tabularx}

\usepackage[normalem]{ulem}
\useunder{\uline}{\ul}{}

\title{LLM and GNN are Complementary: \\ Distilling LLM for Multimodal Graph Learning}

\author{%
  Junjie Xu, Zongyu Wu, Minhua Lin, Xiang Zhang, Suhang Wang \\
  The Pennsylvania State University\\
  \texttt{\{junjiexu, zongyuwu, mfl5681, xzz89, szw494\}@psu.edu}
}

\begin{document}

\maketitle

\begin{abstract}
Recent progress in Graph Neural Networks (GNNs) has greatly enhanced the ability to model complex molecular structures for predicting properties. Nevertheless, molecular data encompasses more than just graph structures, including textual and visual information that GNNs do not handle well. To bridge this gap, we present an innovative framework that utilizes multimodal molecular data to extract insights from Large Language Models (LLMs). We introduce GALLON (Graph Learning from Large Language Model Distillation), a framework that synergizes the capabilities of LLMs and GNNs by distilling multimodal knowledge into a unified Multilayer Perceptron (MLP). This method integrates the rich textual and visual data of molecules with the structural analysis power of GNNs. Extensive experiments reveal that our distilled MLP model notably improves the accuracy and efficiency of molecular property predictions.
\end{abstract}

\section{Introduction}
\label{sec:intro}

In recent years, Graph Neural Networks (GNNs) have demonstrated exceptional prowess in representing learning on graph-structured data~\cite{kipf2017semi, xu2018how, velivckovic2017graph, xu2023shape}. Within the domain of chemistry, GNNs have been notably effective in predicting molecular properties~\cite{gilmer2017neural, wu2018moleculenet}, a task critical for advancements in various domains, such as drug discovery~\cite{gaudelet2021utilizing} and materials science~\cite{reiser2022graph}. By conceptualizing molecules as graphs, GNNs manage to capture the nuanced spatial and chemical relationships that dictate molecular behavior. This unique ability has positioned GNNs as a leading tool in chemistry, enabling more accurate and efficient prediction models that surpass traditional methodologies~\cite{fung2021benchmarking}.

\begin{figure}[t]
    \centering
    \includegraphics[width=0.75\linewidth]{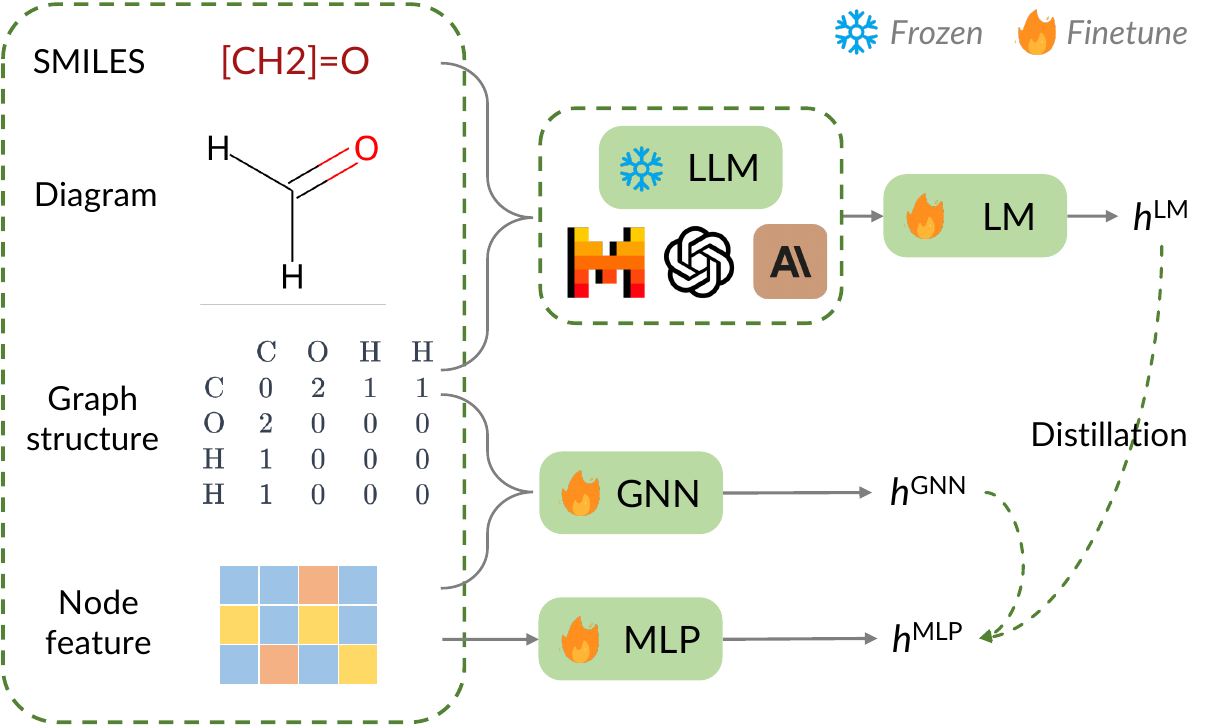}
    \caption{The framework of GALLON (\underline{G}r\underline{a}ph Learning from \underline{L}arge \underline{L}anguage Model Disti\underline{ll}ati\underline{on}).}
    \vskip -1em
    \label{fig:framework}
\end{figure}

Despite the strengths of GNNs in processing graph-structured data, their application scope reveals limitations when encountering molecular data in forms other than graph structures. Molecules, in their essence, exhibit multimodal characteristics that can also be represented in various forms, including textual Simplified Molecular-Input Line-entry System (\textit{SMILES}) strings~\cite{weininger1988smiles} and \textit{visual molecular diagrams}. An example of this multimodal representation is illustrated on the left side of Fig.~\ref{fig:framework}. These modalities provide comprehensive information toward the understanding of molecules. SMILES strings offer a compact and linear textual representation of a molecule's structure, which some textual models can learn. Molecular diagrams provide a visual and intuitive depiction of molecular structures, making specific structures and functional groups, like benzene rings, more identifiable. Additionally, a graph adjacency matrix encapsulates detailed information about the connections between nodes within the molecule. Therefore, each modality provides complementary advantages to each other. However, while GNNs excel in node features and graph structures, they falter in processing and extracting valuable information from other modalities. This limitation underscores a significant gap in the current framework, highlighting the need for models that can understand molecules' modalities from all aspects and learn effective representations.

In parallel with the advancements in GNNs, the evolution of Large Language Models (LLMs)~\cite{radford2019language, brown2020language, yang2023harnessing} and LLMs with vision~\cite{anthropic2024claude, team2023gemini, fu2023challenger, reid2024gemini} has marked a new age in machine learning, characterized by their exceptional skill in parsing and interpreting textual and visual data. For example, GPT4V has shown the ability to answer questions about molecular diagrams~\cite{2023GPT4VisionSC}. However, recent works show that LLMs exhibit a shortcoming in their ability to process graph-structured data~\cite{chen2024exploring, li2023survey}, an area where GNNs excel. Yet, their proficiency in dealing with molecular data represented as SMILES strings and diagrams suggests a valuable complementary role to GNNs. Meanwhile, through the huge amount of training data, LLMs have their understanding of the real world and have prior knowledge about the SMILES string and chemical molecules. Based on these observations, we aim to incorporate the knowledge from LLM and make LLMs and GNNs complementary when dealing with different modalities. One potential way is to fuse LLM and GNN by distilling them into a new model. However, querying LLM is both computationally and financially demanding, making it impractical to apply to each molecule when handling large-scale data.

To this end, we propose to distill LLM and GNN to a smaller Multilayer Perceptron (MLP). On the one hand, the MLP is trained with distillation from both LLM and GNN, incorporating knowledge from the modalities of GNN and LLM where they excel. On the other hand, the MLP offers greater efficiency and cost-effectiveness during the inference process, eliminating the need for querying the LLMs for each single molecule. Fig.~\ref{fig:time_compare} and Fig.~\ref{fig:para_compare} show the time and size efficiency of our model. 

However, several challenges hinder the distillation process. First, while LLMs have demonstrated their power, we empirically find that they struggle with direct molecular property prediction. This difficulty arises not only because such classification or regression tasks challenge text-generation models, but also due to the lack of domain-specific training in LLMs. To mitigate this, we finetune a smaller Language Model (LM) to serve as the encoder for the outputs of the LLM. This approach allows the LM to specialize in the domain relevant to the prediction task, ensuring that the learned representations are more effective for the knowledge distillation process needed for molecular property prediction. Secondly, most methods for graph distillation~\cite{tian2023knowledge, yan2020tinygnn} have concentrated on classification tasks, where the goal is to align the predicted class distribution between the student and teacher models. However, graph regression remains a vital and common task, particularly for predicting molecular properties. Distilling directly from labels in regression tasks presents challenges because the label is a scalar, which provides less information compared to the distribution of labels in classification tasks. Therefore, we employ a mapping function that maps the embeddings of the teacher and student models to a latent space, where we perform the distillation.

In light of these considerations, we propose utilizing multimodal information from molecular data, i.e., SMILES strings, molecular diagrams, graph structures, and node features, to query the Large Language Model (LLM) and obtain a detailed description of the molecule. This description will then serve as the knowledge base for distillation into a smaller MLP model. Furthermore, we propose a novel approach GALLON (\underline{G}r\underline{a}ph Learning from \underline{L}arge \underline{L}anguage Model Disti\underline{ll}ati\underline{on}) that synergizes the advantages of GNNs and LLMs through a distillation process into a Multilayer Perceptron (MLP) model, which makes the inference process more efficient with state-of-the-art performance. By distilling the capabilities of GNNs in handling graph structures and the adeptness of LLMs in processing textual and visual information, we create a unified framework for learning more efficient representations across the multimodality of molecular data. To summarize, our contributions are:
\begin{itemize}[leftmargin=*]
    \item We propose utilizing multimodal molecular data to learn representations and extract prior knowledge from powerful and pretrained large language models by leveraging their multimodality capabilities.
    \item We introduce a framework that synergizes GNNs and LLMs, leveraging their complementary strengths. This framework distills the advantages of GNN and LLM into an MLP, aiming to capture the most effective representations for molecular structures.
    \item We conduct extensive experiments to demonstrate the superiority of our approach in distilling GNN and LLM knowledge into an MLP, which outperforms both GNNs and LLMs across various datasets, achieving greater efficiency and an even smaller model size.
\end{itemize}
\section{Related Work}
\label{sec:related}

\noindent\textbf{LLMs for Graphs}. 
Recent studies have shown significant interest in applying LLMs to graph data, marking a pivotal shift in graph learning research. These models have been adapted to address challenges such as encoding node and edge information and preserving topological structures. TAPE~\cite{he2023harnessing} uses LLMs to generate explanations for nodes, which are then used as augmented features to train GNNs. GraphLLM~\cite{chai2023graphllm} encodes graphs into text for LLM prediction. Das et al.~\cite{das2023modality} explore the integration of graph data with LLMs and the influence of graph multimodalities. CaR~\cite{qian2023can} uses molecular SMILES strings to obtain captions from LLMs, which are then input into another language model for finetuning. Surveys by Li et al.~\cite{li2023survey} and Chen et al.~\cite{chen2024exploring} provide a comprehensive understanding of LLM performance on graphs, dividing them into roles such as enhancer, predictor, and alignment. In contrast to prior works, we use LLMs as teachers in the distillation process, leveraging their prior knowledge about molecules to enhance the learning of the student model.

\noindent\textbf{Graph Knowledge Distillation}. 
Graph Knowledge Distillation constructs a smaller but efficient model by extracting more knowledge from data, aiming for the compressed model and improved performance~\cite{tian2023knowledge}. Various works distill GNNs onto various models. For example, LSP~\cite{yang2020distilling} distills a teacher's GNN to a student's GNN by minimizing the distance of local structure between them. T2GNN~\cite{huo2023t2} employs MLP and GNN as the feature teacher and structure teacher to let a student GNN learn from them. NOSMOG~\cite{tian2022learning} and GLNN~\cite{zhang2021graph} show that distilled MLP also have the ability to outperform GNN methods. Though abundant works adopt different strategies for graph knowledge distillation, most of them do not focus on the molecule data with graph classification tasks, and they cannot deal with the multimodal information in the molecule data.

\noindent\textbf{Multimodal Learning for Molecules}. 
Several studies have focused on leveraging single modalities for molecular representation learning. ChemBERTa~\cite{chithrananda2020chemberta} uses SMILES strings to pretrain a BERT model, treating molecules purely as text. GNN methods~\cite{hu2019strategies, wang2022molecular, wang2019molecule} enhance learning by combining features from individual atoms and their connections. Recent efforts have explored multimodal approaches. MolT5~\cite{edwards2022translation} pretrains on SMILES strings and molecule captions, while MoMu~\cite{su2022molecular} merges molecule graphs with natural language through contrastive learning. MolFM~\cite{luo2023molfm} integrates structures, texts, and knowledge graphs for a comprehensive understanding of molecular properties. In this work, GALLON incorporates SMILES strings, molecular diagrams, and molecule graphs to train effective molecular representations and distill the knowledge to an MLP.

\section{Methodology: GALLON} 
\label{sec:methodology}
In this section, we first define notations and formally define the problem. We then give an overview of GALLON followed by detailed descriptions.


\textbf{Notation and Problem Definition}. Let \(\mathcal{G} = \{\mathcal{V}, \mathcal{E}, \mathbf{X}, \mathcal{S}, \mathcal{I}\}\) be a molecule composed of a node set \(\mathcal{V}\), edge set \(\mathcal{E}\), node features \(\mathbf{X}\), SMILES string \(\mathcal{S}\), and molecular diagram \(\mathcal{I}\). Given a training set \(\mathcal{D}_L = \left\{\mathcal{G}_i\right\}_{i=1}^{|\mathcal{D}_L|}\) with corresponding labels \(\mathcal{Y}_L = \{y_i\}_{i=1}^{|\mathcal{D}_L|}\), and a Large Language Model, we aim to learn a function \(f\) that can predict the labels of unlabelled graphs in \(\mathcal{D}_U = \left\{\mathcal{G}_i\right\}_{i=1}^{|\mathcal{D}_U|}\). 


\noindent\textbf{Overview of GALLON}. An illustration of the framework is shown in Fig.~\ref{fig:framework}, which has three main components: 1) Input the multimodal data into the LLM to extract knowledge from it. 2) Fine-tuning a smaller LM to encode the text outputs of the LLM and pretraining a GNN. 3) Distilling the knowledge extracted from both the GNN and LLM into an MLP. It is important to note a limitation that using MLP with atom input without a graph structure cannot differentiate isomers—compounds with the same formula but different atom arrangements. However, the node features of the molecular dataset include essential structure information (e.g., degree, ring, aromatic, chirality). These structural features enable our method to classify compounds effectively, despite the absence of a graph structure, by leveraging differences in properties such as chirality and ring structure.

\subsection{Extracting Knowledge from LLMs}
\begin{figure}
    \centering
    \includegraphics[width=0.85\linewidth]{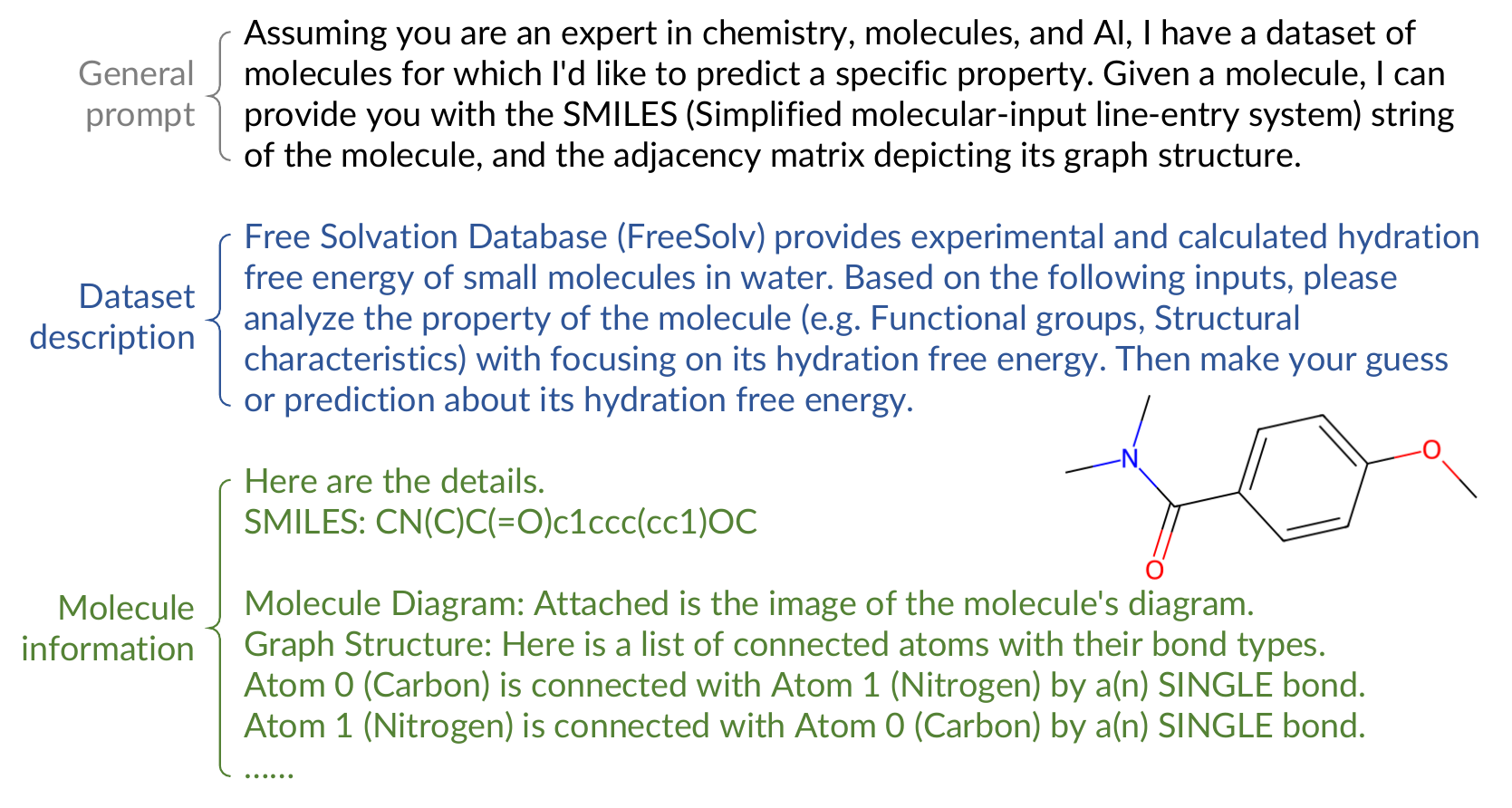}
    \vspace{-0.8em}
    \caption{An example prompt for a molecule of the Freesolv dataset.}
    \vspace{-1.5em}
    \label{fig:prompt}
\end{figure}

\noindent\textbf{Multimodal Information.} As stated previously, LLMs excel at processing text information and are trained on diverse datasets, enabling them to gain extensive knowledge about interpreting molecules and SMILES strings. Consequently, we utilize the SMILES string $\mathcal{S}$ to query the LLM and extract prior knowledge. Furthermore, recent research~\cite{das2023modality} investigates which modalities should be utilized when processing graph data with LLMs. It has been found that images of $k$-hop subgraphs from citation networks can enhance LLM's prediction capabilities on graph data. Unlike citation networks, which are too large to depict in a single image, molecular diagrams can provide a comprehensive view of the whole molecular structure in one image, as illustrated in Fig.~\ref{fig:prompt}. This insight leads us to include molecular diagrams in the multimodal prompts and leverage LLMs with visual capabilities. Lastly, aiming to provide more forms for LLMs to understand the structure, we also feed the graph structure $\mathcal{E}$ of the molecule into the LLM.

\noindent\textbf{Prompt Construction.} To enable the LLM to act like a chemistry expert and generate useful information about the input molecule, it is crucial to design an effective prompt. Fig.~\ref{fig:prompt} shows an example prompt we designed for a molecule in the Freesolv Dataset. A prompt is composed of three parts: (1) The general prompt, which remains the same across all datasets; (2) Dataset-specific description that describes the characteristics of the dataset and the property we want to predict; (3) Molecule information containing the SMILES string, molecular diagram, and graph structure. As shown in Fig.~\ref{fig:prompt}, we convert the graph structure into texts by describing the atom name and bond type for each edge. This designed prompt provides LLM with the general goal, the prediction task, and the multimodal molecule information. Then we use the generated prompt $\mathcal{P}$ to query LLM to get the explanations and descriptions $\mathcal{R}$ of the molecule. For molecule $\mathcal{G}_i$, the process can be formally written as
\begin{equation}
\label{eq:equry_llm}
    \mathcal{R}_i = \text{LLM}(\mathcal{P}_i, \mathcal{E}_i, \mathcal{S}_i, \mathcal{I}_i) ,
\end{equation}
where $\mathcal{E}_i, \mathcal{S}_i, \mathcal{I}_i$ are the graph edges, SMILE string, and molecular diagram respectively.

\subsection{Pretraining and Finetuning}
As the response $\mathcal{R}$ is text-based, we cannot directly distill knowledge from it to MLP. Therefore, we fine-tune a smaller Language Model to serve as an encoder. We input both the original SMILES string $\mathcal{S}$ and the response $\mathcal{R}$ into this Language Model to obtain representations, which are then used for subsequent distillation. We input the SMILES string $\mathcal{S}$ again along with the response $\mathcal{R}$ into the smaller LM to ensure that the primary structural information and interpretive insights are both accurately captured and integrated into the final embedding $h^{\text{LM}}$. Specifically, for molecule $\mathcal{G}_i$ with SMILES string $\mathcal{S}_i$ and the response $\mathcal{R}_i$, the process of obtaining the embedding $h_i^{\text{LM}}$ is written as
\begin{equation}
    h^{\text{LM}}_i = \text{LM}(\mathcal{R}_i, \mathcal{S}_i) \in \mathbb{R}^{H} , 
\end{equation}
where $ H $ is the hidden dimension. We add a transformation function $\mathcal{T}_{\text{LM}}$ on top of the embedding to predict the molecular property as
\begin{equation}
    \hat{y}^{\text{LM}}_i = \mathcal{T}_{\text{LM}} (h^{\text{LM}}_i) \in \mathbb{R}^{d} , 
\end{equation}
where $ d $ denotes the output dimension. For regression tasks, we set $ d=1 $; in classification tasks, $ d $ corresponds to the number of classes. The LM is finetuned based on the labels $ y_i $ with cross-entropy for classification and Root Mean Square Error (RMSE) for regression.

To facilitate the knowledge distillation from graphs, we pretrain a GNN with the graph structure and node features. Specifically, for each molecule $\mathcal{G}_i$, we first adopt GNN with a pooling layer to get graph representation as 
\begin{equation}
    h^{\text{GNN}}_i = \text{GNN}(\mathbf{X}_i, \mathcal{E}_i)  \in \mathbb{R}^{H} ,  
\end{equation}
We then add a transformation layer \( \mathcal{T}_{\text{GNN}} \) to predict the molecular property as
\begin{equation}
    \hat{y}^{\text{GNN}}_i = \mathcal{T}_{\text{GNN}} (h^{\text{GNN}}_i) \in \mathbb{R}^{d} , 
\end{equation} 
Similarly, \( \mathcal{T}_{\text{GNN}} \) is typically implemented using a linear layer for regression, and a linear layer followed by a Softmax function for classification.

\begin{figure}
    \centering
    \begin{minipage}{0.48\linewidth}
        \centering
        \includegraphics[width=0.85\linewidth]{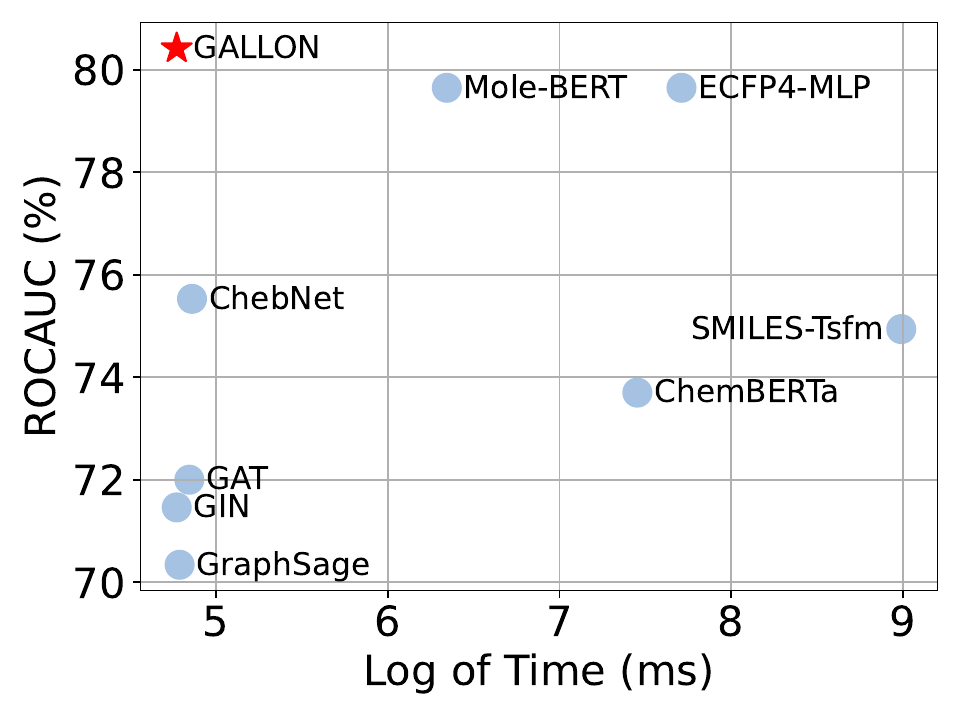}
        \vspace{-1em}
        \caption{ROCAUC vs log of inference time (ms) on the BACE dataset.}
        \label{fig:time_compare}
    \end{minipage}%
    \hfill
    \begin{minipage}{0.48\linewidth}
        \centering
        \includegraphics[width=0.85\linewidth]{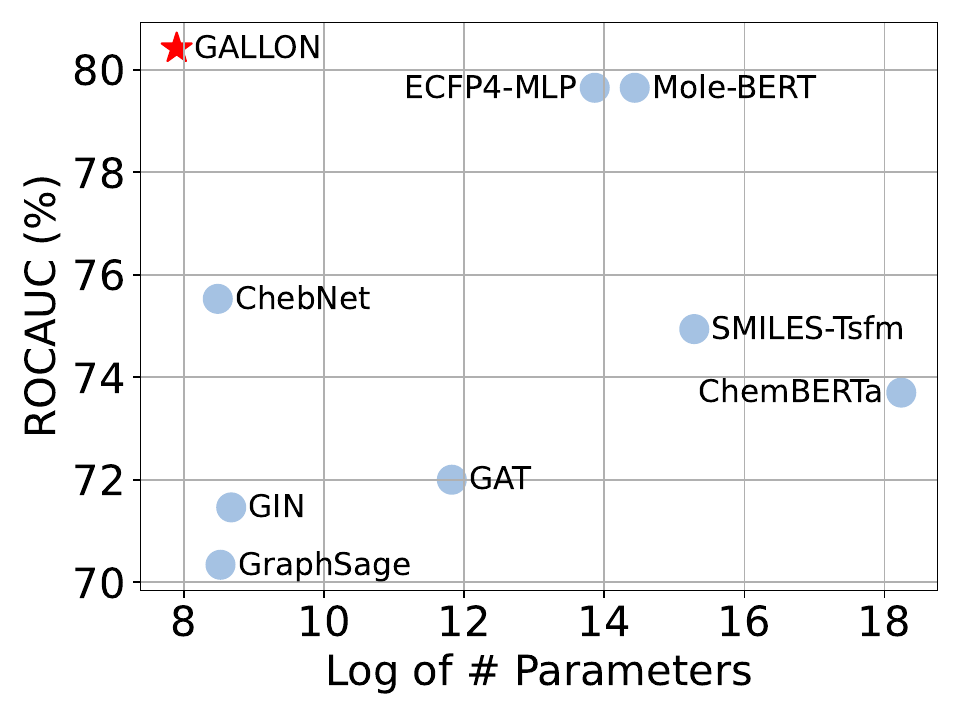}
        \vspace{-1em}
        \caption{ROCAUC vs log of number of parameters on the BACE dataset.}
        \label{fig:para_compare}
    \end{minipage}
\vspace{-1em}
\end{figure}

\subsection{Knowledge Distillation to MLP}
With the learned representations of LLM and GNN, we propose to distill their knowledge into an MLP for several compelling reasons: (\textbf{i}) operating an LLM is both computationally and financially burdensome. In large-scale datasets, querying each molecule through the LLM to extract descriptions and prior knowledge is impractical. Through distillation, we can embed the prior knowledge from the LLM into the MLP, allowing us to use the MLP for efficient inference; and (\textbf{ii}) many studies~\cite{tian2022learning, yang2021extract, zhang2021graph, zheng2021cold} have demonstrated that with effective distillation, an MLP can achieve performance on par with, or even superior to, a GNN, while significantly reducing inference time. Figures~\ref{fig:time_compare} and~\ref{fig:para_compare} demonstrate that GALLON achieves state-of-the-art performance with less time and a smaller model size.

Additionally, the node attributes in the datasets are not simply atomic numbers but include a wide range of information (e.g. chirality, degree, formal charge, hydrogens, radical electrons, hybridization, aromaticity, and ring membership). These comprehensive node features, including some graph structural information, make it feasible to achieve good prediction performance using only the node attributes with an MLP. However, these studies primarily focus on distilling from GNN to MLP and do not incorporate LLM. Therefore, in this section, our goal is to train an MLP via distillation, enabling it to match or exceed the performance of either GNN or LLM alone. Specifically, for a molecule $\mathcal{G}_i$, the MLP takes its node attribute matrix $\mathbf{X}_i$ as input followed by average pooling to get molecule representation as 
\begin{equation}
    Z^{\text{MLP}}_i = \sigma_L(\mathbf{W}_L(\ldots \sigma_1(\mathbf{W}_1\mathbf{X}_i)\ldots))  \in \mathbb{R}^{n_i \times H}, 
    \quad 
    h^{\text{MLP}}_i = \text{Pooling}(Z^{\text{MLP}}_i) \in \mathbb{R}^{H}
\end{equation}
where $n_i$ is the number of atoms in $\mathcal{G}_i$, $L$ is the number of layers, and $\mathbf{W}$ is the weight matrix. We then transform with $\mathcal{T}_{\text{MLP}}$ to get prediction
\begin{equation}
    \hat{y}^{\text{MLP}}_i = \mathcal{T}_{\text{MLP}} (h^{\text{MLP}}_i) \in \mathbb{R}^{d}
\end{equation}
During the training, the loss is calculated between the prediction label and ground truth,
\begin{equation}
    \mathcal{L}_0 = \mathcal{L}_{pred}(\hat{y}^{\text{MLP}}, y) , 
\end{equation}
where $\mathcal{L}_{pred}$ is cross entropy in classification tasks and distance-based RMSE in regression tasks. 

To better distill the knowledge from LLM and GNN, we use two forms of distillation, i.e., representation distillation and label distillation. 

\noindent\textbf{Label Distillation.} Label distillation aims to let the student's prediction mimic the distribution of the teacher's prediction, which can be written as
\begin{equation}
\mathcal{L}^{LD} = \alpha \mathcal{L}_{LM}^{LD} (\hat{y}^{\text{MLP}}, \hat{y}^{\text{LM}}) + \beta \mathcal{L}_{GNN}^{LD} (\hat{y}^{\text{MLP}}, \hat{y}^{\text{GNN}})
\end{equation}
where \(\alpha\) and \(\beta\) serve as hyperparameters that adjust the balance between various distillation losses. Specifically, \(\alpha\) controls the weight of the knowledge distilled from the LLM, while \(\beta\) regulates the influence of the GNN representations. For classification tasks, \(\hat{y}\) are logits in \(\mathbb{R}^C\), where \(C\) represents the number of classes. We employ the KL-Divergence to calculate both \(\mathcal{L}_{LM}\) and \(\mathcal{L}_{GNN}\) for these tasks. Conversely, for regression tasks $\hat{y} \in \mathbb{R}$, we use the RMSE to measure \(\mathcal{L}_{LM}\) and \(\mathcal{L}_{GNN}\).

\noindent\textbf{Representation Distillation.} In regression tasks, the target label is a scalar value, and calculating the distance between two scalar values yields limited information for distillation purposes. To address this challenge, we propose utilizing representation distillation to align the student's learned representations with those of the teacher. Given that LLM and the GNN are trained independently, their respective representations, \(h^{\text{LM}}\) and \(h^{\text{GNN}}\), exist in different feature spaces compared to \(h^{\text{MLP}}\). Directly aligning these representations could be problematic due to the mismatch in feature spaces. To overcome this, we transform them into a common latent space. Subsequently, we facilitate the distillation of knowledge between the student and teachers' models within this shared latent space, which can be mathematically represented as follows:
\begin{equation}
\mathcal{L}^{RD} =
\alpha \mathcal{L}_{LM}^{RD} \left(\mathcal{U}_{\text{MLP}} \left(h^{\text{MLP}}\right), \mathcal{U}_{\text{LM}} \left(h^{\text{LM}}\right)\right) + 
\beta \mathcal{L}_{GNN}^{RD} \left(\mathcal{U}_{\text{MLP}} \left(h^{\text{MLP}}\right), \mathcal{U}_{\text{GNN}} \left(h^{\text{GNN}}\right)\right) , 
\end{equation}
where both $\mathcal{L}_{LM}^{RD}$ and $\mathcal{L}_{GNN}^{RD}$ are implemented by RMSE. For simplicity, in this paper, we adopt label distillation for classification tasks and representation distillation for regression tasks, but both of them can be applied to any task. With the loss function for knowledge distillation, the final objective functions for training MLP for classification and regression are given as
\begin{equation}
    \mathcal{L}^{class} = \mathcal{L}^{0} + \mathcal{L}^{LD} , 
    \qquad
    \mathcal{L}^{reg} = \mathcal{L}^{0} + \mathcal{L}^{RD} . 
\end{equation}

\section{Experiments} 
\label{sec:exp}

In this section, we present experiments to evaluate the effectiveness of the proposed GALLON framework, addressing the following research questions: \textbf{RQ1}: How does the proposed GALLON framework perform compared to GNN and NLP models? \textbf{RQ2}: Do both GNN and LLM contribute to the final results? \textbf{RQ3}: To what extent does multimodality enhance LLM predictions? \textbf{RQ4}: What is the influence of different LLM models on performance?

\subsection{Experimental Setup}

\noindent\textbf{Datasets.} To evaluate the performance of GALLON for molecule property prediction, we adopt seven widely used datasets from MoleculeNet~\cite{wu2018moleculenet}, including four graph classification datasets BACE, BBBP, Clintox, HIV, and three regression datasets ESOL, Freesolv, and Lipophilicity. In this paper, we use ROCAUC for classification and RMSE for regression as evaluation metrics. The detailed description and statistics of the datasets can be found in Appendix~\ref{appen:datasets} and Table~\ref{tab:stat}.

\noindent\textbf{Baselines.}
We compare our model with (i) GNN models, including GCN~\cite{kipf2017semi}, ChebNet~\cite{defferrard2016convolutional}, GraphSAGE~\cite{hamilton2017inductive}, GIN~\cite{xu2018how}, MoleBERT~\cite{xia2023molebert}, and (ii) NLP models, including ECFP4-MLP~\cite{rogers2010extended}, ChemBERTa~\cite{chithrananda2020chemberta}, SMILES-Transformer~\cite{honda2019smiles}. The  descriptions of the baselines are in Appendix~\ref{appen:baselines}.

\noindent\textbf{Settings.}
As suggested by Wu et al.~\cite{wu2018moleculenet}, we adopt scaffold splitting for molecule datasets, which is widely adopted in the molecule domain. There are two kinds of scaffold splitting: \textbf{scaffold splitting} and \textbf{random scaffold splitting}. The difference is introduced in~\ref{appen:settings}. We adopt the ratio of 80\%/10\%/10\% for train/validation/test sets. We use GCN as the GNN backbone and Roberta~\cite{liu2019roberta} as the smaller Language Model (LM) throughout all experiments. For the HIV and Lipophilicity datasets, we query Claude3-Haiku, and for other datasets, we use GPT-4V. We search hyperparameters $\alpha$ and $\beta$ based on the performance on a validation set to achieve optimal results. For a fair comparison, we configured all backbone GCNs and distilled MLPs with 3 layers and a hidden dimension of 32. We conduct the experiments on 10 seeds and report the average performance with standard deviation.

\subsection{Distillation Results}
Table~\ref{tab:scaffold} presents the results of graph classification and regression tasks using scaffold splitting, compared with other GNN and NLP baselines. The boldface indicates the best results, while underlining denotes the second-best. For the classification datasets (BACE, BBBP, Clintox, and HIV), the ROCAUC score (\%) is shown; for the regression datasets (ESOL, FreeSolv, and Lipo), the RMSE is shown. Our proposed GALLON framework achieves state-of-the-art performance on BACE, BBBP, Clintox, and FreeSolv, and performs comparably to the best baselines on HIV, ESOL, and Lipo with a simpler MLP structure. We attribute this to the distilled MLP leveraging the strengths of both MLP and GNN, utilizing their complementary aspects—specifically, the prior knowledge embedded in LLMs and the capability of learning from graph structures in GNNs. This synergistic combination enhances the overall performance.

\begin{table}[]
\centering
\caption{Prediction performance with scaffold splitting.}
\label{tab:scaffold}
\resizebox{\textwidth}{!}{%
\begin{tabular}{c|cccc|ccc}
\toprule
            & \textbf{BACE↑}      & \textbf{BBBP↑}      & \textbf{Clintox↑}         & \textbf{HIV↑}       & \textbf{ESOL↓}     & \textbf{Freesolv↓} & \textbf{Lipo↓}     \\ \hline
GCN         & 73.47±3.78          & 63.99±1.70          & 69.22±2.42                & 70.19±2.17          & 1.29±0.03          & 3.33±0.40          & 0.89±0.02          \\
ChebNet     & 75.53±1.60          & 67.07±0.75          & 68.95±2.81                & 73.78±1.66          & 1.32±0.04          & 3.27±0.17          & 1.04±0.05          \\
GraphSage   & 72.00±6.05          & 66.56±2.93          & 84.01±3.56                & 74.81±0.99          & 1.32±0.09          & 3.12±0.30          & 0.87±0.01          \\
GIN         & 70.34±2.57          & 62.53±3.27          & 73.40±3.73                & 72.54±3.32          & 1.26±0.08          & 3.71±0.79          & {\ul 0.84±0.01}    \\
Mole-BERT   & 71.46±4.74          & 67.52±1.17          & 66.46±7.26                & {\ul 75.42±1.13} & 1.39±0.07          & 3.98±0.62          & 0.78±0.01          \\
ECFP4-MLP   & {\ul 79.65±1.88}    & 61.84±0.37          & 70.03±1.57                & 69.58±0.61          & 1.67±0.30          & 4.07±1.21          & \textbf{0.83±0.02} \\
ChemBERTa   & 73.70±2.39          & {\ul 70.44±1.20}    & 97.41±1.64                & \textbf{76.60±1.14} & 1.77±0.04          & 3.97±0.13          & 1.19±0.10          \\
SMILES-Tsfm & 74.94±1.22          & 68.21±0.53          & {\ul 98.43±0.46} & 72.00±0.80          & \textbf{1.06±0.08} & {\ul 2.74±0.94}    & 2.74±0.94          \\
GALLON      & \textbf{80.42±1.06} & \textbf{72.38±0.63} & \textbf{99.15±0.57}       & 75.39±0.40          & {\ul 1.18±0.02}    & \textbf{2.03±0.09} & 0.90±0.01          \\ 
\bottomrule
\end{tabular}%
}
\vspace{-1em}
\end{table}

\subsection{Contributions of LLM and GNN}
In this section, we aim to explore the influence of the teachers' models. We conduct distillation using GNN or LLM solely as the teacher model, as well as using the combination of LLM+GNN as teacher models. The performance results are shown in Table~\ref{tab:ablation:scaffold} and Table~\ref{tab:ablation:random_scaffold} under different split settings. GNN$\rightarrow$MLP denotes distilling solely from GNN to MLP, while (GNN+LLM)$\rightarrow$MLP represents the original GALLON framework, distilling from both GNN and LLM to MLP. The results reveal the following observations: (1) Across the seven datasets, the distilled MLP outperforms both single GNN, LLM, or MLP, demonstrating that MLP can leverage the unique strengths of each model. This suggests that the MLP benefits from the graph structural information of the GNN and the contextual knowledge of the LLM, leading to a more effective learning process. (2) The (GNN+LLM)$\rightarrow$MLP configuration achieves the best performance, indicating that the synergy between GNN and LLM contributes significantly to the final results.

\begin{table}[]
\centering
\caption{The comparison between different distillation settings with \textbf{scaffold} splitting.}
\label{tab:ablation:scaffold}
\small
\resizebox{\textwidth}{!}{%
\begin{tabular}{ccccccc}
\toprule
\textbf{Dataset} & \textbf{GNN} & \textbf{LLM} & \textbf{(GNN+LLM)$\rightarrow$MLP} & \textbf{GNN$\rightarrow$MLP} & \textbf{LLM$\rightarrow$MLP} & \textbf{MLP}\\
\hline
BACE             & 73.47±3.78   & 73.57±2.67   & 80.42±1.06         & 77.93±1.38         & 78.50±2.00   & 73.11±2.55     \\
BBBP             & 63.99±1.70   & 71.73±1.93   & 72.38±0.63         & 67.24±0.36         & 66.92±0.61   & 60.21±0.64     \\
Clintox          & 69.22±2.42   & 99.06±0.64   & 99.15±0.57         & 83.94±4.51         & 84.99±1.51   & 70.89±4.93     \\
HIV              & 70.19±2.17   & 76.60±0.61   & 75.39±0.40         & 70.94±2.47         & 71.14±0.64   & 66.17±0.75     \\
ESOL             & 1.29±0.03    & 2.24±0.19    & 1.18±0.02          & 1.22±0.02          & 1.23±0.02    & 1.29±0.03      \\
Freesolv         & 3.33±0.40    & 4.20±0.09    & 2.03±0.09          & 2.15±0.12          & 2.16±0.05    & 2.59±0.06      \\
Lipo             & 0.89±0.02    & 1.16±0.06    & 0.90±0.01          & 0.92±0.01          & 0.91±0.01    & 0.96±0.01      \\
\bottomrule
\end{tabular}}
\vspace{-2em}
\end{table}

\begin{table}[]
\centering
\caption{The comparison between different distillation settings with \textbf{random scaffold} splitting.}
\label{tab:ablation:random_scaffold}
\small
\resizebox{\textwidth}{!}{%
\begin{tabular}{ccccccc}
\toprule
\textbf{Dataset} & \textbf{GNN} & \textbf{LLM} & \textbf{(GNN+LLM)$\rightarrow$MLP} & \textbf{GNN$\rightarrow$MLP} & \textbf{LLM$\rightarrow$MLP} & \textbf{MLP} \\
\hline
BACE     & 79.08±6.61 & 83.87±3.47  & 80.34±2.68          & 76.84±1.69        & 79.50±2.96      & 70.73±3.02     \\
BBBP     & 81.23±6.42 & 80.97±1.94  & 84.84±1.24          & 83.21±1.58        & 83.78±1.42      & 79.98±2.31     \\
Clintox  & 81.61±6.04 & 72.08±11.55 & 93.71±4.35          & 90.98±4.98        & 91.20±6.88      & 87.39±4.69     \\
HIV      & 72.33±4.73 & 74.80±4.04  & 74.00±2.26          & 72.63±3.60        & 72.69±3.71      & 67.66±3.35     \\
ESOL     & 1.28±0.11  & 1.81±0.24   & 1.21±0.09           & 1.28±0.06         & 1.24±0.09       & 1.51±0.13      \\
Freesolv & 3.18±0.70  & 4.66±1.23   & 1.85±0.40           & 2.14±0.40         & 2.03±0.43       & 2.87±0.71      \\
Lipo     & 0.89±0.06  & 1.18±0.10   & 0.89±0.05           & 0.90±0.05         & 0.90±0.05       & 0.95±0.05      \\
\bottomrule
\end{tabular}
}
\end{table}

\subsection{Efficiency Comparison}
In this experiment, we evaluate the performance of various models on the BACE dataset by comparing their ROCAUC scores against the log of inference time and the log of the number of parameters. Each result is conducted on the whole dataset 100 times to calculate the averaged inference time. Fig.~\ref{fig:time_compare} and Fig.~\ref{fig:para_compare} illustrates the relationship between ROCAUC and the log of inference time and model size, showing that GALLON achieves the best performance with the fastest inference time and smallest model size. These results reveal that our approach effectively distills knowledge from GNN and LLM into MLP, leading to state-of-the-art performance in terms of both efficiency and accuracy.

\subsection{Influence of Multimodality} 
\begin{figure}
\vspace{-2em}
    \centering
    \begin{minipage}[b]{0.5\linewidth}
        \centering
        \includegraphics[width=0.7\linewidth]{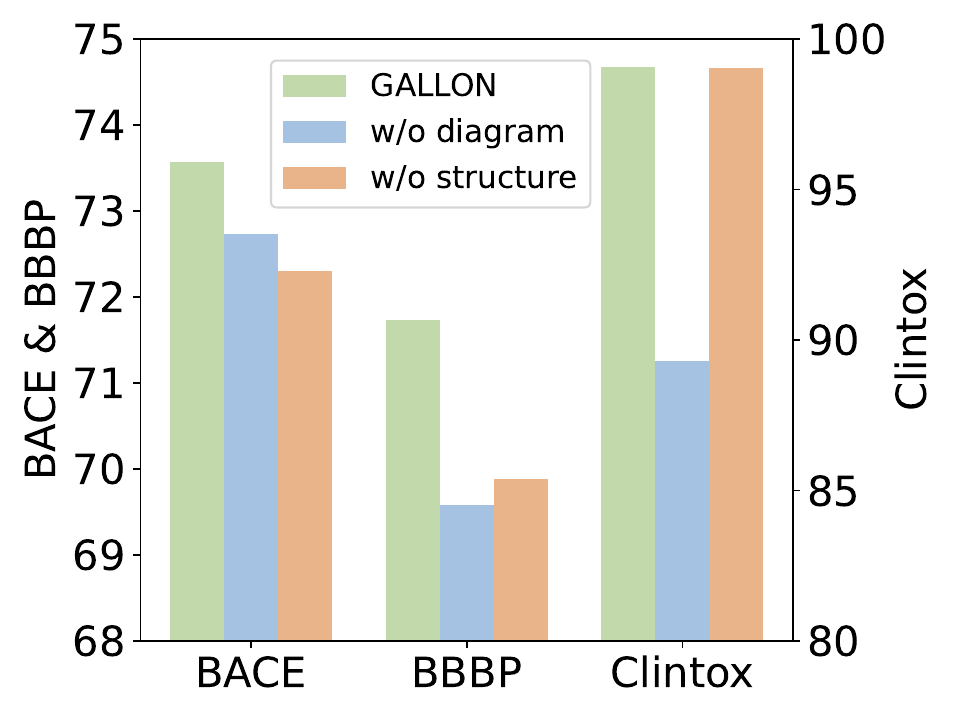}
        \vspace{-0.8em}
        \caption{Ablation study of multimodalities.}
        \label{fig:influence_multimodality}
    \end{minipage}%
    \begin{minipage}[b]{0.5\linewidth}
        \centering
        \includegraphics[width=0.58\linewidth]{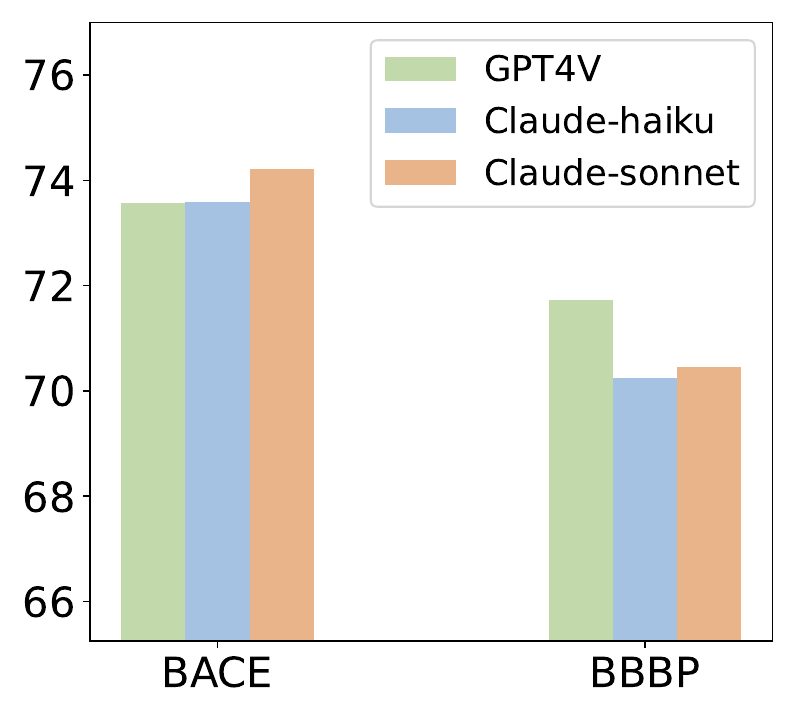}
        \vspace{-0.8em}
        \caption{Performance of different LLMs.}
        \label{fig:influence_model}
    \end{minipage}
\end{figure}

In the GALLON framework, we incorporate all of SMILES string, molecular diagram, and graph structure when querying the LLM. To explore the influence of each multimodality, we ablate the molecular diagram and graph structure respectively and plot the results of prediction of the finetuned LM in Fig.~\ref{fig:influence_multimodality}. We conduct the experiments on BACE, BBBP, and Clintox, where all results are based on GPT4V and scaffold splitting. We find that the original GALLON with all the multimodality achieves the best performance, which means each modality contribute to extracting richer knowledge from LLM. To be more specific, removing molecular diagrams results in significant performance drops, due to the lack of overview of the molecule, such as some functional groups. Similarly, excluding graph structures leads to declines in BACE and BBBP, as the graph structure is easier for LLM to understand atomic connectivity insights. These findings underscore the importance of leveraging multiple modalities to enhance molecular representation and prediction accuracy.

\subsection{Influence of Large Language Model}
Fig.~\ref{fig:influence_model} illustrates the performance of different LLMs (GPT4V, Claude3-haiku, and Claude3-sonnet) on BACE and BBBP datasets with scaffold splitting. The results show that each dataset has its own best-performing LLM, although the differences are not significant. (1) On BACE, Claude3-sonnet slightly outperforms GPT4V and Claude3-haiku, suggesting it may capture the chemical properties relevant to this dataset more effectively. By contrast, on BBBP dataset, GPT4V achieves the highest performance. (2) Claude3-sonnet consistently outperforms Claude3-haiku, due to its larger model size and capacity.

\section{Case Study} 
\label{sec:case}
In this section, we explore how multimodality aids in extracting prior knowledge from LLMs. We use a molecule from BACE dataset as an example. Its SMILES string is 
\[
\resizebox{\textwidth}{!}{\text{S1(=O)C[C@@H](Cc2cc(OC(C(F)(F)F)C(F)(F)F)c(N)c(F)c2)[C@H](O)[C@@H]([NH2+]Cc2cc(ccc2)C(C)(C)C)C1}}
\]
and its diagram is shown in Fig.~\ref{fig:case_diagram}. We query GPT-4V using three different multimodal combinations: (1) SMILES string, diagram, and graph structure, (2) SMILES string and diagram, (3) SMILES string and graph structure, and (4) SMILES string only. We present each response in Appendix~\ref{appen:response}.

Through the responses, we summarize the knowledge that LLMs provide with various multimodal combinations in Table 6. The table shows how many functional groups and structural characteristics the LLM can identify from the input modality and whether it can facilitate interactions with BACE-1 and predict the binding result. Based on the table, we make the following observations:
(1) Inputs of all SMILES strings, diagram, and graph structure can extract the most comprehensive information from the LLM. This indicates that both the diagram and graph structure contribute significantly to the overall analysis.
(2) SMILES + Diagram can identify more functional groups than SMILES + Graph Structure. This suggests that the molecular diagram provides a better way to identify certain functional groups due to its intuitive and comprehensive overview of the molecule.
(3) Although SMILES + Graph Structure has a similar number of identified items as SMILES only, the responses show that SMILES only lacks detailed explanations for each functional group. This undermines the richness of the extracted knowledge.
(4) Only the combination of SMILES, diagram, and graph structure provide the most abundant structural characteristics. This emphasizes the importance of multimodal inputs for a complete structural analysis.
(5) All combinations can identify interactions relevant to binding with BACE-1. However, predictions about the molecule’s activity are more reliably made with the combination of SMILES, diagram, and graph structure, as well as SMILES and diagram, compared to other combinations.

\begin{table}[h]
\vspace{-0.5em}
\centering
\caption{The analysis of information provided by various multimodal combinations on an example molecule of BACE dataset.}
\label{tab:case}
\resizebox{0.9\textwidth}{!}{%
\begin{tabular}{cc|cccc}
\toprule
\multicolumn{2}{c|}{Multimodalities} &
  \begin{tabular}[c]{@{}c@{}}SMILES\\ Diagram\\ Graph Structure\end{tabular} &
  \begin{tabular}[c]{@{}c@{}}SMILES\\ Diagram\end{tabular} &
  \begin{tabular}[c]{@{}c@{}}SMILES\\ Graph Structure\end{tabular} &
  SMILES \\ \hline
\multicolumn{1}{c|}{\multirow{9}{*}{\begin{tabular}[c]{@{}c@{}}Functional\\ Groups\end{tabular}}}          & Sulfonamide Group    & $\checkmark$ & $\checkmark$ & $\checkmark$ & $\checkmark$ \\
\multicolumn{1}{c|}{}                                                                                      & Aromatic Ring        & $\checkmark$ & $\checkmark$ & $\checkmark$ & $\checkmark$ \\
\multicolumn{1}{c|}{}                                                                                      & Hydroxyl Group       & $\checkmark$ & $\checkmark$ &   &   \\
\multicolumn{1}{c|}{}                                                                                      & Amine Group          & $\checkmark$ & $\checkmark$ & $\checkmark$ & $\checkmark$ \\
\multicolumn{1}{c|}{}                                                                                      & Ether Group          & $\checkmark$ & $\checkmark$ &   &   \\
\multicolumn{1}{c|}{}                                                                                      & Fluorine Atom        & $\checkmark$ & $\checkmark$ & $\checkmark$ & $\checkmark$ \\
\multicolumn{1}{c|}{}                                                                                      & Amide Bond           & $\checkmark$ &   &   &   \\
\multicolumn{1}{c|}{}                                                                                      & Thioether Group      &   & $\checkmark$ &   &   \\
\multicolumn{1}{c|}{}                                                                                      & Tertiary Butyl Group &   &   & $\checkmark$ & $\checkmark$ \\ \hline
\multicolumn{1}{c|}{\multirow{3}{*}{\begin{tabular}[c]{@{}c@{}}Structural\\ Characteristics\end{tabular}}} & Rigid \& Flexible    & $\checkmark$ &   &   &   \\
\multicolumn{1}{c|}{}                                                                                      & Chiral Centers       & $\checkmark$ & $\checkmark$ &   & $\checkmark$ \\
\multicolumn{1}{c|}{}                                                                                      & Molecular Bulk       & $\checkmark$ &   & $\checkmark$ &   \\ \hline
\multicolumn{2}{c|}{Interactions}                                                                                                 & $\checkmark$ & $\checkmark$ & $\checkmark$ & $\checkmark$ \\ \hline
\multicolumn{2}{c|}{Prediction}                                                                                                   & $\checkmark$ & $\checkmark$ & $\checkmark$ & $\checkmark$ \\ \bottomrule
\end{tabular}%
}

\end{table}

\section{Conclusion and Future Work}
In this paper, we introduced the GALLON framework, a novel approach for molecular property prediction that leverages the complementary strengths of Graph Neural Networks and Large Language Models. By distilling knowledge from both GNNs and LLMs into a more efficient Multilayer Perceptron model, we addressed the limitations of each individual modality and enhanced the overall predictive performance. One of the limitations of the paper is the distillation method. Future research can build upon this work in several directions. First, investigating advanced distillation techniques and more sophisticated mapping functions may improve the efficiency and effectiveness of the distillation process. Second, exploring other multimodal data representations and their integration into the GALLON framework could further enhance prediction accuracy and generalizability. Lastly, extending the framework to other domains beyond molecules, such as materials science and biology, could validate its broader applicability and utility.

\newpage
\bibliography{ref}
\bibliographystyle{plain}
\appendix
\newpage

\section{Experimental Details}
\label{appen:exp}

\subsection{Dataset Details}
\label{appen:datasets}

\begin{itemize}[leftmargin=*]
    \item \textbf{BACE}: This dataset focuses on inhibitors of human beta-secretase 1. It includes both quantitative (IC50 values) and qualitative (binary labels) binding results. It includes 1,513 compounds with their 2D structures and binary labels for the classification task. 
    \item \textbf{BBBP}: The Blood-brain barrier penetration (BBBP) dataset comes from a study focused on modeling and predicting the permeability of the blood-brain barrier. This contains 2,050 compounds with 2D structures and binary labels indicating whether a compound can penetrate the blood-brain barrier (BBB) or not. 
    \item \textbf{Clintox}: This dataset compares drugs approved by the FDA and drugs that have failed clinical trials for toxicity reasons. It includes two classification tasks for 1,484 drug compounds with known chemical structures: (i) clinical trial toxicity (or absence of toxicity) and (ii) FDA approval status.
    \item \textbf{HIV}: This dataset is introduced by the Drug Therapeutics Program (DTP) AIDS Antiviral Screen. It tests the ability to inhibit HIV replication for 41,127 compounds with binary labels indicating whether a compound is activate or not. 
    \item \textbf{ESOL}: The ESOL (Estimated Solubility) dataset is a small-scale molecule dataset focuses on the water solubility of various compounds for a regression task. It includes quantitative solubility data for 1,128 compounds, with their 2D structures and measured solubility values in mols per liter. 
    \item \textbf{Freesolv}: This provides data on the hydration-free energies of small molecules. It includes both experimental and calculated hydration-free energies for 642 compounds, along with their 2D structures.
    \item \textbf{Lipophilicity}: This dataset contains data on the octanol/water partition coefficient (logP) values for various compounds. It includes 4,200 compounds with their 2D structures and experimental results of octanol/water distribution coefficient (logD at pH 7.4). 
\end{itemize}

\begin{table}[h]
\centering
\caption{The statistics and tasks of datasets.}
\label{tab:stat}
\resizebox{\textwidth}{!}{%
\begin{tabular}{cccccccc}
\toprule
    & \#Graphs & Avg. \#nodes & Avg. \#edges  & \#Features & \#Classes & Task  & Metric         \\ 
         \hline
BACE     & 1,513     & 34.1         & 73.7  & 9 & 1       & Classification      & ROCAUC \\
BBBP     & 2,050     & 23.9         & 51.6  & 9 & 1       & Classification      & ROCAUC  \\
Clintox  & 1,484     & 26.1         & 55.5  & 9 & 2       & Classification      & ROCAUC  \\
HIV      & 41,127    & 25.5         & 54.9  & 9 & 1       & Classification      & ROCAUC  \\
ESOL     & 1,128     & 13.3         & 27.4  & 9 & -       & Regression          & RMSE      \\
Freesolv & 642       & 8.7          & 16.8  & 9 & -       & Regression          & RMSE      \\
Lipo     & 4,200     & 27.0         & 59.0  & 9 & -       & Regression          & RMSE      \\ 
\bottomrule
\end{tabular}}
\end{table}

\subsection{Baselines}
\label{appen:baselines}

\begin{itemize}[leftmargin=*]
    \item \textbf{GCN}~\cite{kipf2017semi}: Graph Convolutional Network (GCN) is one of the most popular MPNNs using 1-hop neighbors in each layer.
    \item \textbf{ChebNet}~\cite{defferrard2016convolutional}: ChebNet uses Chebyshev polynomial to approximate the filter function. It is a more generalized form of GCN.
    \item \textbf{GraphSAGE}~\cite{hamilton2017inductive}: Graph Sample and Aggregation (GraphSAGE) is a scalable and inductive framework that leverages node feature information by sampling and aggregating features from a fixed-size set of neighbors. 
    \item \textbf{GIN}~\cite{xu2018how}: Graph Isomorphism Network (GIN) is a powerful variant of MPNNs designed to achieve maximum expressive power for graph representation learning, effectively distinguishing different graph structures through injective node aggregations.
    \item \textbf{MoleBERT}~\cite{xia2023molebert}: MoleBERT is a self-supervised learning strategy for pretraining GNNs, specifically designed for molecules. It introduces two novel pretraining tasks: Masked Atoms Modeling (MAM) at the node level and triplet masked contrastive learning (TMCL) at the graph level. MAM involves randomly masking some discrete codes and then pretraining GNNs to predict them. TMCL models the heterogeneous semantic similarity between molecules for effective molecule retrieval. A variant of VQ-VAE\cite{van2017neural} is proposed as a context-aware tokenizer to encode atom attributes into chemically meaningful discrete codes.
    \item \textbf{ECFP4-MLP}~\cite{rogers2010extended}: Extended-connectivity fingerprint (ECFP) is a manually constructed fingerprinting method specifically developed to identify molecular characteristics. 4 means quantifying substructures with diameters of up to 4. After getting fingerprinting from ECFP4, we train a multilayer perceptron.
    \item \textbf{ChemBERTa}~\cite{chithrananda2020chemberta}: ChemBERTa is based on the Roberta~\cite{liu2019roberta}. It consists of 12 attention heads and 6 layers, resulting in 72 distinct attention mechanisms. It is then pre-trained on a dataset of 77M unique SMILES from PubChem~\cite{kim2019pubchem}.
    \item \textbf{SMILES-Transformer}~\cite{honda2019smiles}: SMILES Transformer has 4 Transformer blocks. Each Transformer block has 4 attention heads with 256 embedding dimensions and includes two linear layers. It is pre-trained on a dataset consisting of SMILES randomly selected from ChEMBL24.
\end{itemize}

\subsection{Settings}
\label{appen:settings}

\noindent\textbf{Models.} In this paper, we employ GCN~\cite{kipf2017semi} as the GNN backbone. To ensure fair comparisons, all backbone GCNs and distilled MLPs are configured with 3 layers and a hidden dimension of 32. Roberta~\cite{liu2019roberta} is used as the smaller Language Model (LM) for all experiments. For the HIV and Lipophilicity datasets, we use Claude3-Haiku, and for the other datasets, we utilize GPT-4V.

\noindent\textbf{Splitting.} Scaffold splitting divides a dataset based on chemical scaffolds in a deterministic manner, ensuring structural diversity across train, validation, and test sets, thereby enhancing the model's ability to generalize to new scaffolds. Random scaffold splitting also partitions the dataset based on scaffolds but introduces randomness in the assignment, achieving a balance between structural diversity and randomized subset distribution.

\noindent\textbf{Training.} We utilize an 80\%/10\%/10\% split for the train, validation, and test sets. To ensure fair comparisons, the best hyperparameter configurations for each method are selected using the validation set, and we report the mean accuracy and variance across 10 different seeds on the test set. The hyperparameters \(\alpha\) and \(\beta\) are searched within the set \{0, 0.001, 0.005, 0.01, 0.05, 0.1, 0.5, 1.0, 5.0, 10.0\}.

\section{Prompts}
\label{appen:prompt}

Here we show the dataset description part of the prompt used for each dataset. 

\begin{table}[h]
\centering
\caption{The description part of the prompts for each dataset.}
\label{tab:dataset_description}
\begin{tabular}{ll}
\toprule
\textbf{Dataset} & \textbf{Description} \\ \hline
BACE &
  \begin{tabular}[c]{@{}l@{}}The BACE dataset provides and binary label binding results for a set of inhibitors \\ of human $\beta$-secretase 1 (BACE-1). Based on the following inputs, please analyze \\ the property of the molecule (e.g. Functional groups, Structural characteristics) and \\ analyze the binding results for a set of inhibitors of human beta-secretase (BACE-1)?\end{tabular} \\ \hline
BBBP &
  \begin{tabular}[c]{@{}l@{}}As a membrane separating circulating blood and brain extracellular ﬂuid the \\ blood-brain barrier blocks most drugs, hormones, and neurotransmitters. Based on \\ these inputs, please analyze the property of the molecule (e.g. Functional groups, \\ Structural characteristics) and analyze if the molecule is permeable to the \\ blood-brain barrier?"\end{tabular} \\ \hline
Clintox &
  \begin{tabular}[c]{@{}l@{}}Could you analyze the given molecule based on the provided inputs and detail the \\ factors influencing its potential for clinical trial toxicity or non-toxicity? \\ Additionally, please assess factors that might impact its FDA approval status.\end{tabular} \\ \hline
HIV &
  \begin{tabular}[c]{@{}l@{}}The HIV dataset tests the ability to inhibit HIV replication for over 40,000 compounds. \\ Based on the following inputs, please analyze the property of the molecule \\ (e.g. Functional groups, Structural characteristics) with focusing on its ability to inhibit\\  HIV replication. Then make your guess or prediction (active or inactive).\end{tabular} \\ \hline
ESOL &
  \begin{tabular}[c]{@{}l@{}}Based on these inputs, please analyze the property of the molecule \\ (e.g. Functional groups, Structural characteristics), and which properties of the \\ molecule can affect its water solubility? Also try to guess its solubility.\end{tabular} \\ \hline
Freesolv &
  \begin{tabular}[c]{@{}l@{}}Free Solvation Database (FreeSolv) provides experimental and calculated hydration \\ free energy of small molecules in water. Based on the following inputs, please analyze \\ the property of the molecule (e.g. Functional groups, Structural characteristics) with \\ focusing on its hydration free energy. Then make your guess or prediction about its \\ hydration free energy.\end{tabular} \\ \hline
Lipo &
  \begin{tabular}[c]{@{}l@{}}Lipophilicity is an important feature of drug molecules that affects both membrane \\ permeability and solubility. Based on the following inputs, please analyze the molecule \\ and give some details of factors that can affect octanol/water distribution coefficient \\ (logD at pH 7.4). Then make your guess or prediction about its lipophilicity. \end{tabular}  \\
\bottomrule
\end{tabular}
\end{table}


\section{More Details of Case Study}
\label{appen:response}

\begin{figure}[ht]
    \centering
    \includegraphics[width=0.5\textwidth]{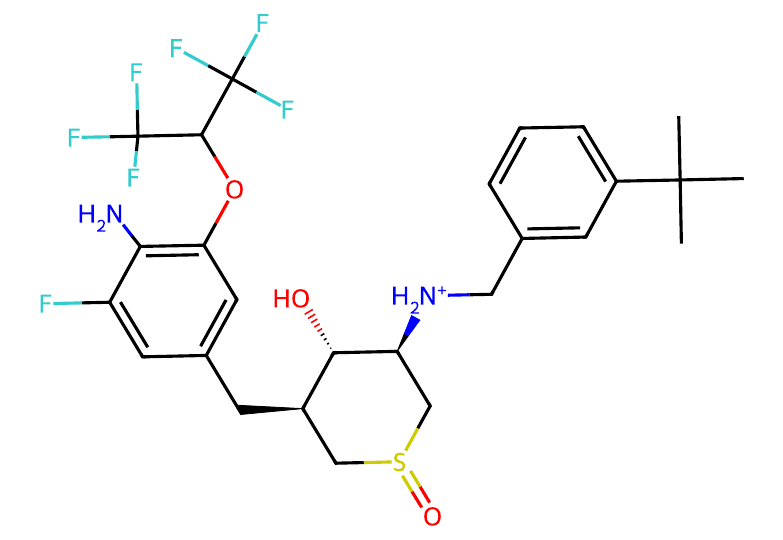}
    \caption{The diagram of the example molecule.}
    \label{fig:case_diagram}
\end{figure}

This section presents the responses we obtained from the LLM for the example molecule in Case Study~\ref{sec:case}. These texts are used to analyze the richness of information provided in Table~\ref{tab:case}. Fig.~\ref{fig:case_diagram} displays the diagram of the example molecule.

\begin{table}[ht]
\centering
\caption{The response of LLM using SMILES string, diagram, and graph structure.}
\label{tab:case_response1}
\begin{tabular}{|>{\raggedright\arraybackslash}p{0.95\textwidth}|}
\hline
(1) SMILES string, diagram, graph structure \\
\hline
The provided SMILES string, molecule diagram, and graph structure describe a complex organic molecule with various functional groups and structural features that may be relevant for its activity as an inhibitor of $\beta$-secretase 1 (BACE-1).\\

Analyzing the molecule based on the SMILES and the provided image:\\

1. \textbf{Functional Groups:}\\
- Sulfonamide Group (Sulfur connected to Oxygen by double bond and Nitrogen by single bond): This is often found in BACE-1 inhibitors as it can form hydrogen bonds with the enzyme.\\
- Aromatic Rings: There are multiple phenyl groups (benzene rings), which are important for $\pi$-$\pi$ stacking interactions with aromatic residues in proteins.\\
- Hydroxyl Groups (-OH): These are capable of forming hydrogen bonds, which can be key for binding to the active site.\\
- Amines (both primary with NH2+ and tertiary with the nitrogen having three carbon substituents): These can engage in hydrogen bonding and ionic interactions, especially the positively charged amine, which can interact with negatively charged residues in the active site.\\
- Ether Groups (Oxygen connected to Carbon by single bonds): These can provide some flexibility and help position other functional groups within the binding pocket.\\
- Fluorine Atoms: The presence of multiple fluorine atoms suggests the molecule is designed to increase its lipophilicity for membrane penetration or create specific interactions with the enzyme due to the high electronegativity of fluorine.\\
- Amide Bonds: The carbon-nitrogen bonds in the context of carbonyl groups (C=O) indicate amide linkages, which can also participate in hydrogen bonding.\\

2. \textbf{Structural Characteristics:}\\
- The presence of both rigid (aromatic rings) and flexible (alkyl chains) elements in the molecule can help it to adapt its conformation to fit the binding pocket of the enzyme.\\
- The molecule has chiral centers, which are indicated by the symbols "@@" in the SMILES string, affecting how the molecule may interact with the chiral environment of the enzyme's active site.\\
- There is significant molecular bulk near the periphery of the molecule, due to the presence of tert-butyl groups and trifluoromethyl groups, which could be important for binding affinity and specificity.\\

In terms of the BACE-1 inhibition, the functional groups such as the sulfonyl group, aromatic rings, and amine groups would be essential for forming interactions with the active site of the enzyme. The molecule is likely designed to fit within the binding pocket of BACE-1, forming hydrogen bonds, hydrophobic interactions, and possibly ionic interactions with key amino acid residues.\\

The success of this compound as a BACE-1 inhibitor may be dependent on these interactions, as well as its overall shape and the ability to adopt a conformation that is complementary to the enzyme binding site.\\

To predict binding results or to further analyze the structural features and their potential interaction with BACE-1, computational techniques such as molecular docking or molecular dynamics could be applied if more details about the active site and its interactions with ligands are available. Additionally, experimental assays would be required to confirm the inhibitory activity of the molecule.\\
\hline
\end{tabular}
\end{table}

\begin{table}[h!]
\centering
\caption{The response of LLM using SMILES string, and diagram.}
\label{tab:case_response2}
\begin{tabular}{|>{\raggedright\arraybackslash}p{0.95\textwidth}|}
\hline
(2) SMILES string and diagram \\
\hline
Based on the SMILES string provided and the diagram attached, we can infer several properties of the molecule and discuss its potential interactions as an inhibitor of human beta-secretase 1 (BACE-1), focusing on functional groups and structural characteristics relevant to its activity. \\ 

First, examining the structural features from the SMILES string and the diagram: \\ 
1. Heterocyclic Components: There appear to be several heterocyclic components, one of which is likely to contribute to the molecule's ability to interact with the active site of BACE-1. \\ 
2. Fluorinated Substituents: The presence of multiple fluorine atoms could influence the binding affinity and specificity due to their electronegative nature, and they are often found in inhibitors to enhance metabolic stability and bioavailability. \\ 
3. Sulfonamide Group: The S(=O)(=O)N motif represents a sulfonamide group, which is a common feature in many drug molecules that can contribute to hydrogen bonding with the enzyme's active site. \\ 
4. Chiral Centers: The molecule contains multiple chiral centers, as indicated by the "[C@@H]" and "[C@H]" annotations in the SMILES string. The configuration of these centers can significantly influence the binding affinity and selectivity of the molecule for BACE-1. \\ 
5. Amine Group: The molecule contains a positively charged amine group "[NH2+]", which suggests it may form salt bridges or hydrogen bonds with BACE-1. \\ 
6. Aromatic Rings: There are aromatic rings, which can facilitate $\pi-\pi$ stacking interactions with aromatic residues in the active site of the enzyme. \\ 
7. Hydroxyl Group: There is a hydroxyl group, a polar functional group that can form hydrogen bonds with BACE-1. \\ 
8. Ethers and Thioethers: Both ether (C-O-C) and thioether (C-S-C) groups are present, which may contribute to the binding through interactions with nearby residues. \\ 

When assessing binding results for BACE-1 inhibitors, a few types of interactions can be analyzed: \\

- Hydrogen Bonds: Strong H-bond donors and acceptors in the molecule can form specific interactions with the enzyme's active site, stabilizing the inhibitor-enzyme complex. \\ 
- Hydrophobic Interactions: Non-polar regions of the inhibitor may engage in hydrophobic contacts with non-polar pockets of the active site, increasing affinity. \\ 
- Electrostatic Interactions: Charged groups, like the amine, can form electrostatic interactions with oppositely charged residues of the enzyme or with metal ions that may be present in the active site. \\ 
- Conformational Fit: The overall shape of the molecule must fit well into the active site of BACE-1 to exhibit high binding affinity. \\ 

Without experimental data, predicting the exact mode of inhibition is not feasible. However, the presence of these functional groups and structural features suggests that the molecule is designed to fit into the enzyme's active site and have the necessary interactions for inhibition. \\ 

In silico methods, such as molecular docking and dynamics simulations, coupled with in vitro assays, would be employed to predict the binding affinity and conformation of the inhibitor in the BACE-1 active site to more accurately assess its potential as a therapeutic molecule. \\ 
\hline
\end{tabular}
\end{table}

\begin{table}[h!]
\centering
\caption{The response of LLM using SMILES string, and graph structure.}
\label{tab:case_response3}
\begin{tabular}{|>{\raggedright\arraybackslash}p{0.95\textwidth}|}
\hline
(3) SMILES string and graph structure \\
\hline
Based on the provided SMILES string and graph structure, the molecule has several noteworthy features: \\

1. Sulfonamide group: The molecule contains a sulfonamide group (S(=O)(=O)N), which is known to be an essential pharmacophore in drug design and can form hydrogen bonds with the target protein, increasing the binding affinity. \\ 
2. Fluorinated aromatic ring: The molecule has a fluorinated phenyl ring, which can enhance the metabolic stability and lipophilicity of the compound, potentially improving its bioavailability and membrane permeability. \\ 
3. Secondary amine: The presence of a secondary amine (NH2) might be involved in hydrogen bonding interactions with the enzyme's active site, which can be a key factor for the inhibitory activity. \\ 
4. Tertiary butyl group: The bulky tert-butyl group (C(C)(C)C) may contribute to the molecule's steric hindrance, affecting its binding conformation and possibly improving selectivity for BACE-1 by fitting into specific hydrophobic pockets. \\ 
5. Multiple aromatic rings: The presence of several interconnected aromatic rings provides a rigid framework that can favorably stack with the aromatic amino acid residues in the BACE-1 active site, enhancing the $\pi-\pi$ interactions. \\ 

Upon analyzing the binding results for this inhibitor against human beta-secretase (BACE-1), the mentioned functional groups and structural characteristics suggest that the molecule may exhibit strong binding affinity and potency. The sulfonamide group can form key hydrogen bonds, the fluorinated ring can improve pharmacokinetic properties, and the bulky groups may enhance selectivity. However, the actual binding results would require empirical data obtained from biochemical assays, molecular docking studies, or computational simulations. \\ 

Using AI-based predictive models, one could input the molecule's structural representation, such as the SMILES string or molecular graph, and obtain predictions on its binding affinity, selectivity, and potential inhibitory activity against BACE-1. Such models are trained on large datasets of known inhibitors and their corresponding biological activities, allowing for the rapid virtual screening of novel compounds. \\
\hline
\end{tabular}
\end{table}

\begin{table}[h!]
\centering
\caption{The response of LLM using SMILES string only.}
\label{tab:case_response4}
\begin{tabular}{|>{\raggedright\arraybackslash}p{0.95\textwidth}|}
\hline
(4) SMILES string only \\
\hline
Based on the SMILES string provided, the molecule appears to have a complex structure with several functional groups. The molecule contains a sulfone group (S(=O)), an amine group ([NH2+]), multiple fluorine atoms (F), and an aromatic ring (cc). Additionally, the molecule also has chiral centers, indicated by the "@" symbol, which can impact its binding affinity to the target enzyme. \\ 

The presence of fluorine atoms in the molecule suggests that it may have strong binding interactions with the BACE-1 enzyme, as fluorine is known to form strong hydrogen bonds with proteins. The aromatic ring and amine group may also contribute to the binding affinity through pi stacking interactions and hydrogen bonding, respectively. \\ 

The molecule also contains a methoxy group (OC) and a tert-butyl group (C(C)(C)C), which could impact its overall shape and steric hindrance, potentially affecting its binding to the BACE-1 enzyme. \\ 

To analyze the binding results for this molecule as an inhibitor of BACE-1, we would need to take into account the specific interactions between the molecule and the enzyme's active site. This could be done using molecular docking simulations or other computational chemistry techniques, which would allow us to predict the binding affinity and identify key interactions contributing to the inhibition of the enzyme. \\ 

Overall, based on the structural characteristics of the molecule, it appears to have several features that could contribute to strong binding interactions with BACE-1, making it a potential inhibitor of the enzyme. However, further analysis using computational tools or experimental techniques would be needed to confirm its efficacy as a BACE-1 inhibitor. \\ 
\hline
\end{tabular}
\end{table}

\end{document}